\documentclass[10pt,twocolumn,letterpaper]{article}

\usepackage{cvpr}              

\usepackage[dvipsnames]{xcolor}

\definecolor{cvprblue}{rgb}{0.21,0.49,0.74}
\usepackage[pagebackref,breaklinks,colorlinks,citecolor=cvprblue]{hyperref}
\usepackage{soul}
\usepackage{multirow}

\def\paperID{*****} 
\def\confName{CVPR}
\def\confYear{2024}

\title{CLIP-Embed-KD: Computationally Efficient Knowledge Distillation Using Embeddings as Teachers}

\author{Lakshmi Nair\\
Georgia Institute of Technology\\
Atlanta, GA, USA\\
}

\begin{document}
\maketitle
\begin{abstract}
\textit{Contrastive Language-Image Pre-training} (\textit{CLIP}) has been shown to improve zero-shot generalization capabilities of language and vision models. In this paper, we extend CLIP for efficient knowledge distillation, by utilizing embeddings as teachers. Typical knowledge distillation frameworks require running forward passes through a teacher model, which is often prohibitive in the case of billion or trillion parameter teachers. In these cases, using only the embeddings of the teacher models to guide the distillation can yield significant computational savings. Our preliminary findings show that CLIP-based knowledge distillation with embeddings can outperform full scale knowledge distillation using $9\times$ less memory and $8 \times$ less training time.
\end{abstract} 
\section{Introduction}
\label{sec:intro}

\begin{figure}
  \centering
  \begin{tabular}{@{}c@{}}
    \includegraphics[width=.47\textwidth]{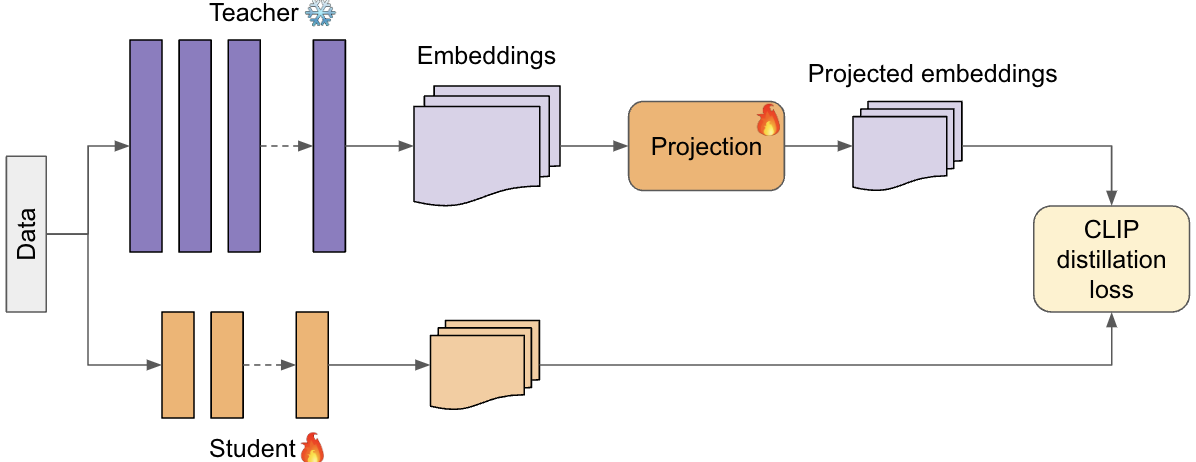} \\[\abovecaptionskip]
    \small (a) CLIP-Teacher-KD
  \end{tabular}

  \vspace{\floatsep}

  \begin{tabular}{@{}c@{}}
    \includegraphics[width=.47\textwidth]{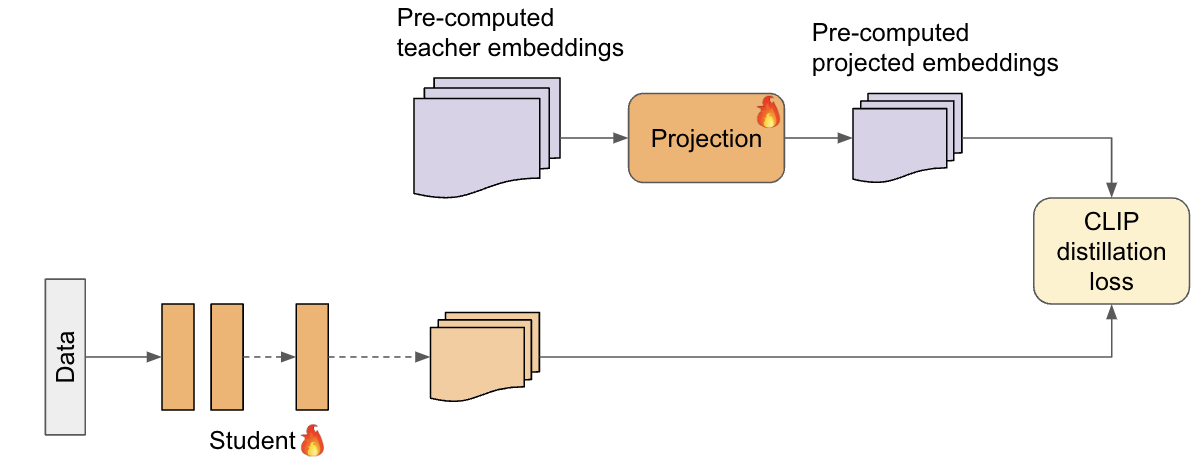} \\[\abovecaptionskip]
    \small (b) CLIP-Embed-KD
  \end{tabular}

  \caption{\textbf{CLIP-Teacher-KD} uses the teacher model to generate embeddings for every training sample. \textbf{CLIP-Embed-KD} uses pre-computed teacher embeddings thus avoiding the need to run forward passes through the teacher model for every sample.}
  \label{fig:summary_fig}
\end{figure}

Contrastive Language-Image Pre-training (CLIP) \cite{radford2021learning}, introduced in 2021, uses natural language supervision to efficiently learn visual concepts. CLIP gained significant popularity owing to its improved zero-shot transfer robustness compared to regularly pre-trained ImageNet models. CLIP involves pre-training an image encoder and text encoder to predict image-text pairings within a dataset. At its core, CLIP uses a constrastive objective function that first computes the scaled pair-wise cosine similarity between image embeddings (of an image encoder) and text embeddings (of a text encoder) to generate output logits. During training, the logits are compared to labels that match input images to their corresponding ground truth text embeddings, using a cross entropy loss. Aligning the image and text modalities in this way enables CLIP models to achieve robust zero-shot performances across classification tasks. Originally applied to visual classification tasks on images, several follow up papers have since successfully extended the application of CLIP to other data modalities such as videos \cite{wasim2023vita}, point clouds \cite{zhang2022pointclip}, and multimodal contexts \cite{ali2023clip}.

In this brief paper, we investigate the application of CLIP to knowledge distillation (KD). Knowledge distillation is the process of transferring (or \textit{distilling}) the knowledge of a larger teacher model into a smaller, more compressed, student model \cite{hinton2015distilling}. This is achieved by comparing the outputs of the teacher and student models using an appropriate loss function (called \textit{distillation loss}). Existing knowledge distillation approaches can be categorized as response-based, feature-based, or instance-based \cite{gou2021knowledge}. Response-based KD compares outputs of the final classification layer (i.e., logits) of the teacher and student models for each training sample. Feature-based KD compares the outputs of the final and intermediate layers (i.e., feature maps) of the student and teacher models. Lastly, instance-based KD compares the similarity between intermediate feature maps of pairs of layers in the teacher and student models. However, all distillation approaches require performing several forward passes through both the teacher and student models for comparing the corresponding outputs. \ul{This can be prohibitive when the teacher models are billion parameters in size as with recent models} \cite{dehghani2023scaling}, \ul{and we wish to run the KD training on limited computational resources (e.g., mobile devices)}.

Intuitively, aligning the teacher and student feature maps can be likened to the alignment of text and image embeddings with CLIP. Prior work on knowledge distillation with CLIP uses CLIP-trained vision and text encoder models, with token selection and confidence weighting applied to a distillation loss that is formulated as the L1 difference between the base model tokens and teacher tokens \cite{wang2022clip}. Other approaches have applied knowledge distillation on models that are pre-trained with CLIP, for performing video-language retrieval task using KL-divergence loss \cite{pei2023clipping}. In \cite{yang2023clip}, the authors compare CLIP-based knowledge distillation approaches through different loss functions including mean-squared error (MSE) or KL-divergence. In contrast to existing methods, we explore the question: \textbf{\textit{Can pre-computed embeddings obtained from the teacher model be used to train the student model in knowledge distillation?}} We investigate two methods in this context.

\begin{itemize}
    \item \textbf{CLIP-Teacher-KD}: We investigate the direct application of CLIP's contrastive pre-training objective in the distillation loss of a teacher and student, along with the student's typical training loss (similar to feature-based KD).
    \item \textbf{CLIP-Embed-KD}: We extend the above to investigate if pre-computed teacher embeddings can be used in place of the teacher model in order to effectively align the student. This eliminates the need to run repeated forward passes of the teacher, except for the initial pre-computation step.
\end{itemize}

\ul{Our goal with this preliminary work is not to present state-of-the-art KD results, rather to present initial findings on the potential of extending CLIP with teacher embeddings for more \textit{computationally efficient} knowledge distillation.}

\section{Approach}
\label{sec:approach}

\subsection{Knowledge distillation}
Knowledge distillation (KD) involves three components: a) teacher model which is typically a large model exhibiting the behavior that we would like to distill into a smaller model; b) the smaller student model; c) a distillation loss that captures the difference between the teacher and student outputs with the goal of minimizing their differences:
\begin{align}
    \mathcal{L_D}(z_s, z_t, \hat{y}) = \alpha_1 \mathcal{L_{CE}}(z_s, \hat{y}) + \alpha_2 \mathcal{L_{KD}}(z_s, z_t)
\end{align}
Here, $\mathcal{L_D}$ is the overall distillation loss over the output logits of the student, $z_s$ and the output logits of the teacher, $z_t$. The term $\mathcal{L_{CE}}$ is the typical cross-entropy supervision loss of the student logits and ground truth labels $\hat{y}$. Additionally, $\mathcal{L_{KD}}$ represents the distillation term that captures the difference between teacher and student logits, and is commonly the KL-divergence between $z_s$ and $z_t$. The scalars $\alpha_1$, $\alpha_2$ represent a weighting of the two losses. Minimizing $\mathcal{L_D}$ aligns the student and teacher model behaviors while minimizing the student loss with respect to the ground truth $\hat{y}$. Follow up research has proposed several variations on the loss functions \cite{gou2021knowledge}. However, in all cases of KD, even if the teacher parameters are frozen (i.e., no backward pass is needed), computing the logits $z_t$ for the loss function requires forward passes through the teacher model which is often much larger than the student model \cite{gou2021knowledge}. Thus, a significant proportion of the computational resources in KD are allocated towards the operation of the teacher. 

Our preliminary work is motivated to address this issue and we present our approach in two parts: CLIP-Teacher-KD evaluates the performance of using CLIP's pre-training objective within distillation, but requires running repeated forward passes through the teacher model. We subsequently adapt this to CLIP-Embed-KD which uses pre-computed embeddings in place of the full teacher model.


\subsection{CLIP-Teacher-KD}

In CLIP-Teacher-KD, we first pass the input training data through the student and teacher models and extract the output embeddings for the [CLS] token. The [CLS] token (often used in models like BERT \cite{devlin2018bert} and ViT \cite{dosovitskiy2020image}) represents a summarized feature map which is extracted and used for the final classification. The [CLS] teacher embeddings are projected through a learnable projection layer into the embedding dimensions of the student model. We further normalize the resultant embeddings and compute the dot product of the normalized teacher and student embeddings. We pass the resultant dot product with ground truth labels into a cross entropy loss ($\mathcal{L}_{clip}$). We use an identity matrix as the ground truth (since teacher embedding $i$ should correspond to student embedding $i$). We also generate the final output logits of the student model through its classifier, which consumes the [CLS] token embeddings and maps them to class probabilities using softmax. The distillation loss ($\mathcal{L_D}$) is a combination of the cross entropy loss of the student logits and ground truth labels, and $\mathcal{L}_{clip}$:
\begin{gather}
    E_s = f_{student}(x); E_t = f_{teacher}(x);  \\
    \hat{E}_t = E_t \cdot W^T_{proj} \\
    \mathcal{L}_{clip} = \mathcal{L_{CE}}(\lVert E_s \rVert \cdot \lVert \hat{E}_t \rVert ^T, G) \\
    \mathcal{L_{D}}(E_s, \hat{E}_t, z_s, \hat{y}) = \alpha_1\mathcal{L_{CE}}(z_s, \hat{y}) + \alpha_2\mathcal{L}_{clip}(E_s, \hat{E}_t)
\end{gather}
Here, $E_s \in \mathbb{R}^{B \times D_s}$ and $E_t \in \mathbb{R}^{B \times D_t}$ denote student and teacher embeddings of the [CLS] token, respectively. $B$ denotes the batch size and $G$ is the identity matrix $\mathbb{I}^{B \times B}$. The learnable projection $W_{proj} \in \mathbb{R}^{D_s \times D_t}$ projects teacher embeddings into the student's embedding dimension. We use $\alpha_1 = 0.5$ and $\alpha_2 = (1 - \alpha_1) = 0.5$ in our experiments.

\subsection{CLIP-Embed-KD}
In this approach, we adapt CLIP-Teacher-KD to use teacher embeddings in place of the full teacher model, to improve the computational efficiency of knowledge distillation by avoiding the need to run repeated forward passes through the teacher model for each incoming training sample. We begin by randomly sampling $N$ samples of data for each class $c$ in the dataset. We obtain teacher embeddings of the [CLS] tokens for the collected data and compute a cumulative representation of each class embedding, by simply averaging the embeddings of each class along the embedding dimension. This gives an \textit{``averaged''} embedding for each class representing the teacher embeddings $E_t$. In our future work, we seek to explore alternate ways of computing more faithful representations of these teacher embeddings.  


The averaged embeddings are $E_t \in \mathbb{R}^{N_c \times D_t}$, where $N_c$ is number of classes. As with CLIP-Teacher-KD, we project the averaged embeddings using a learnable projection layer to the embedding dimensions of the student. When computing $\mathcal{L}_{clip}$ with the averaged embeddings, the ground truth is a $B \times N_c$ matrix, representing a one-hot encoding of the labels for each sample in the batch. That is, each row of the matrix is a one-hot encoding of the label for the corresponding sample. The dot product of $E_s$ and the transposed averaged teacher embeddings result in an output of dimensions $B \times N_c$. The output is passed into a cross entropy loss function with the ground truth matrix to compute $\mathcal{L}_{clip}$. The remainder of the loss function is same as in Equation (5). We make our code publicly available at: \url{https://github.com/lnairGT/CLIP-Distillation/}.

\begin{figure}[!t]
\centering
\includegraphics[width=0.47\textwidth]{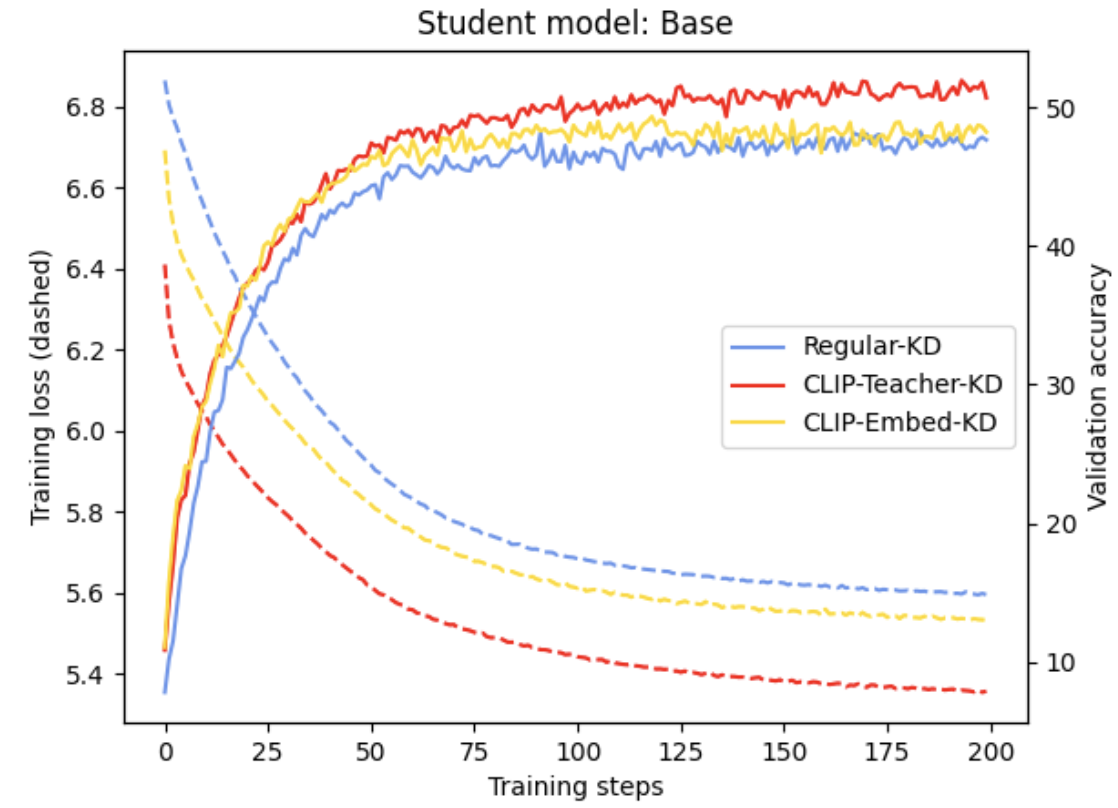}
\caption{Training and validation curves of base student model.}
\label{fig:small-model-training}
\end{figure}

\begin{figure}[!t]
\centering
\includegraphics[width=0.47\textwidth]{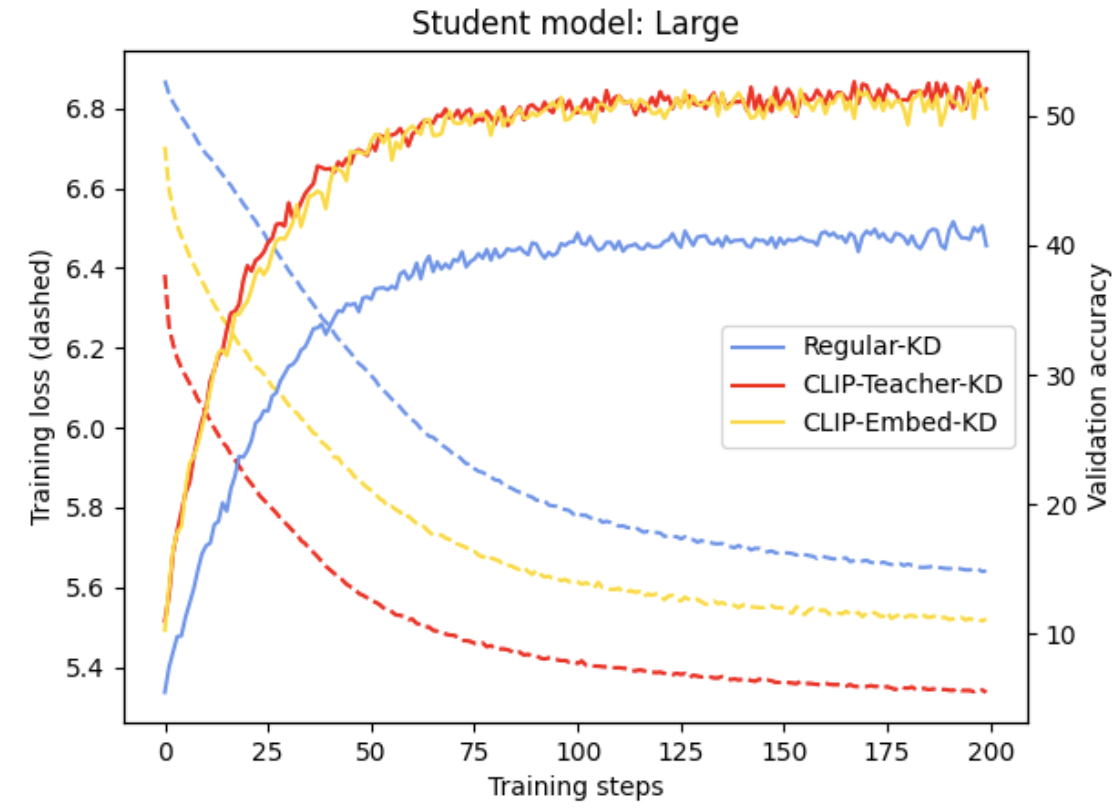}
\caption{Training and validation curves of large student model.}
\label{fig:large-model-training}
\end{figure}
\section{Experiments and Results}
\label{sec:experiments}

\begin{table}[]
\centering
\begin{tabular}{c|c|c|c|c}
\textbf{Student} & \textbf{Layers} & \textbf{Embed dim} & \textbf{Heads} & \multicolumn{1}{l}{\textbf{MLP}} \\ \hline
Base & 6 & 256 & 8 & 1024 \\
Large & 10 & 512 & 8 & 2048 \\ \hline
\textbf{Teacher} & \textbf{Layers} & \textbf{Embed dim} & \textbf{Heads} & \multicolumn{1}{l}{\textbf{MLP}} \\ \hline
Base-16/32 & 12 & 768 & 12 & 3072 \\
Large-16/32 & 24 & 1024 & 16 & 4096
\end{tabular}
\caption{Architecture specification of the models.}
\label{tab:student_arch}
\end{table}

In this section, we evaluate CLIP-Teacher-KD and CLIP-Embed-KD on image classification with CIFAR100. Our goal is not to achieve state-of-the-art performances, rather to investigate the value of using CLIP loss in knowledge distillation through the following research questions: 

\noindent a) \textit{How does using the CLIP distillation loss from Equation (5) compare to the regular KD loss in Equation (1)?}

\noindent b) \textit{How does CLIP-Embed-KD compare to CLIP-Teacher-KD in terms of computational efficiency and accuracy?}

We use Vision Transformers (ViT) \cite{dosovitskiy2020image} and the teacher architectures (Table \ref{tab:student_arch}) use ImageNet pre-trained checkpoints available on HuggingFace, with variants of \textit{google/vit-base-patch16-224-in21k}. We use two teacher and student sizes: base and large. The teacher models use patch sizes $16$, $32$ and student models use patch size $4$. We use batch size of $64$, and image sizes $32 \times 32$ and $64 \times 64$. The teacher models always use an image size of $224 \times 224$ to generate their outputs (logits or embeddings). For CLIP-Embed-KD we compute the averaged embeddings over 100 samples per class. We train all student models for 200 epochs with a learning rate of 0.0001, cosine annealing and weight decay of 0.1.


\begin{table}[]
\centering
\begin{tabular}{c|c|c}
\textbf{Teacher} & \textbf{CLIP-Embed-KD} & \textbf{CLIP-Teacher-KD} \\ \hline
Base-16 & 49.32 & 51.68 \\
Large-16 & 49.67 & 51.92 \\ \hline
Base-32 & 49.34 ($17\times\downarrow$ mem) & 52.61 \\
Large-32 & \textbf{50.33} ($59\times\downarrow$ mem) & 51.95
\end{tabular}
\caption{Performance on base student with image $32 \times 32$.}
\label{tab:regular_vs_cluster}
\end{table}

\begin{table}[]
\centering
\begin{tabular}{c|c|c|c}
\textbf{Student} & \textbf{Teacher} & \textbf{Image sz} & \textbf{CLIP-Embed-KD} \\ \hline
Base & & 64 & 53.28 \\
Large & Large-16 & 32 & 52.92 \\
Large & & 64 & \textbf{54.36} \\ \hline
Base & & 64 & 53.40 \\
Large & Large-32 & 32 & 52.77 \\
Large & & 64 & \textbf{54.92}
\end{tabular}
\caption{Scaling CLIP-Embed-KD to larger models and images.}
\label{tab:larger_model_distill}
\end{table}

\begin{figure}[!t]
\centering
\includegraphics[width=0.475\textwidth]{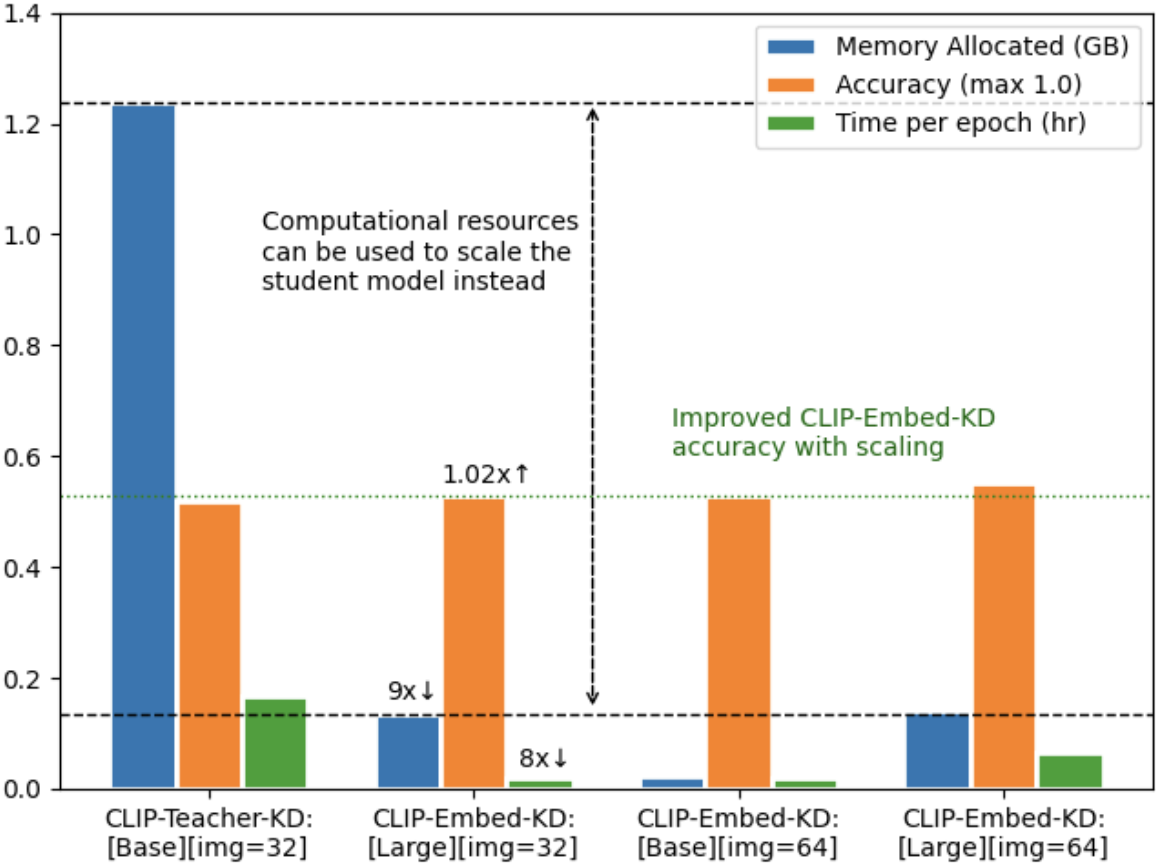}
\caption{Training resource utilization (with Large-32 teacher) of CLIP-Embed-KD, CLIP-Teacher-KD [student-size][image-size]: CLIP-Embed-KD scales well (to larger models and larger images) to outperform CLIP-Teacher-KD for much less memory.}
\label{fig:resource-compare}
\end{figure}

\begin{figure}[t!]
\centering
\includegraphics[width=0.47\textwidth]{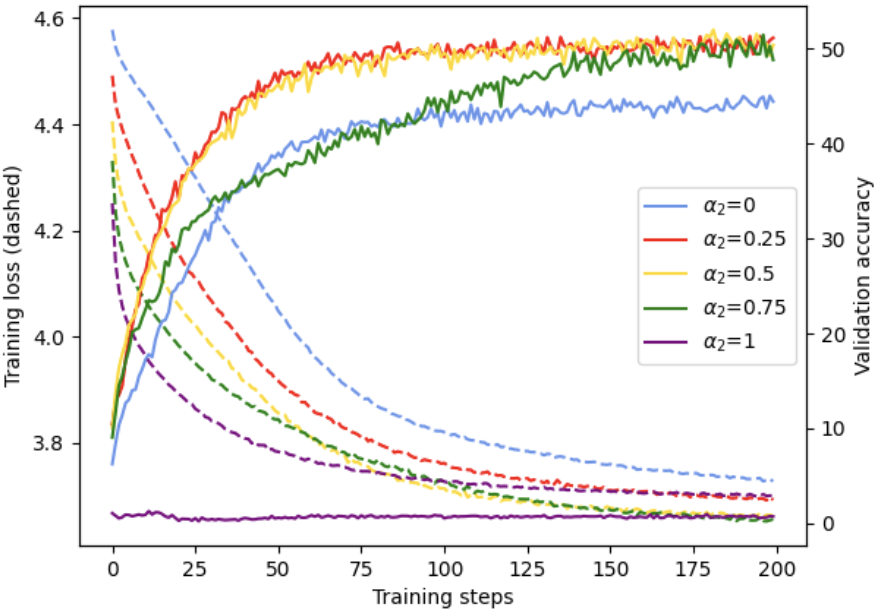}
\caption{Accuracy vs. $\alpha_2$ ($\alpha_1 = 1 - \alpha_2$). Non-zero weighting of both losses yield the best performance ($\alpha_2 = 1$ causes overfitting).}
\label{fig:val-compare-scaling}
\end{figure}

\subsection{CLIP distillation loss vs. regular KD loss}
Figures \ref{fig:small-model-training} and \ref{fig:large-model-training} compare the training loss and validation accuracy curves of regular KD that uses equation (1), with CLIP-Teacher-KD and CLIP-Embed-KD on small and large student models with a fixed ViT-base teacher of patch size 16. Here, we see that using \ul{CLIP's contrastive objective outperforms the regular KD loss function from equation (1)}. CLIP-Teacher-KD slightly outperforms CLIP-Embed-KD. CLIP-Teacher-KD and CLIP-Embed-KD are much closer in performance for the larger student model (Figure \ref{fig:large-model-training}). We further analyze the differences between the two approaches.

\subsection{CLIP-Teacher-KD vs. CLIP-Embed-KD}
Table \ref{tab:regular_vs_cluster} compares the accuracy of CLIP-Teacher-KD vs. CLIP-Embed-KD for different teacher sizes and patch sizes, for the base student model. We note that for CLIP-Embed-KD, final accuracy of the student seems to have a small improvement when using the large teacher models over the base ones. This pattern is not explicit with CLIP-Teacher-KD where the base teacher model has a slightly improved student accuracy compared to the large teacher at patch size 32. Since CLIP-Embed-KD uses pre-computed teacher embeddings to train the student, larger teacher sizes potentially contribute to improved quality of the average embeddings.

\smallskip \ul{\textit{CLIP-Embed-KD is computationally more resource efficient than CLIP-Teacher-KD, and can scale better to outperform CLIP-Teacher-KD.}} In Table \ref{tab:regular_vs_cluster}, CLIP-Embed-KD achieves roughly about $\approx$2\% lesser accuracy than CLIP-Teacher-KD, since the averaged embeddings result in some loss of information per sample compared to CLIP-Teacher-KD. \ul{However, CLIP-Teacher-KD uses more memory as teacher size grows, whereas, CLIP-Embed-KD uses fixed memory for embeddings alone, leading to better scaling behavior ($17\times$, $59 \times$ less memory used)}. In Figure \ref{fig:resource-compare}, we see that CLIP-Embed-KD can get higher accuracy with larger student models and larger image sizes \textit{while staying at a significantly lower computational budget} in comparison to CLIP-Teacher-KD. Eliminating the need to store the teacher model and run repeated forward passes through it, allows the freed resources to be better utilized for training larger student models instead. Even with slightly larger students, the memory used is much lesser than the teacher, and CLIP-Embed-KD outperforms CLIP-Teacher-KD in accuracy.

Validation accuracy for different $\alpha_2$ values are in Figure \ref{fig:val-compare-scaling} (for large student; large-32 teacher; and image size 32). We see that non-zero weighting of both losses yield the best performance. Using only $\mathcal{L}_{clip}$ (i.e., $\alpha_2 = 1$) leads to overfitting. Using only the typical supervised cross-entropy loss (i.e., $\alpha_2 = 0$) performs worse than when $\mathcal{L}_{clip}$ is included.


\section{Future Work}
\label{sec:conclusion}
Our preliminary work demonstrates the potential of CLIP-Embed-KD for computationally efficient distillation. In the future, we seek to evaluate alternative teacher embedding representations that can be used in CLIP-Embed-KD to further improve the performance, and evaluate our approach on billion-trillion parameter NLP models and diverse datasets. 



{
    \small
    \bibliographystyle{ieeenat_fullname}
    \bibliography{main}

\begin{thebibliography}{12}
\providecommand{\natexlab}[1]{#1}
\providecommand{\url}[1]{\texttt{#1}}
\expandafter\ifx\csname urlstyle\endcsname\relax
  \providecommand{\doi}[1]{doi: #1}\else
  \providecommand{\doi}{doi: \begingroup \urlstyle{rm}\Url}\fi

\bibitem[Ali and Khan(2023)]{ali2023clip}
Muhammad Ali and Salman Khan.
\newblock Clip-decoder: Zeroshot multilabel classification using multimodal clip aligned representations.
\newblock In \emph{Proceedings of the IEEE/CVF International Conference on Computer Vision}, pages 4675--4679, 2023.

\bibitem[Dehghani et~al.(2023)Dehghani, Djolonga, Mustafa, Padlewski, Heek, Gilmer, Steiner, Caron, Geirhos, Alabdulmohsin, et~al.]{dehghani2023scaling}
Mostafa Dehghani, Josip Djolonga, Basil Mustafa, Piotr Padlewski, Jonathan Heek, Justin Gilmer, Andreas~Peter Steiner, Mathilde Caron, Robert Geirhos, Ibrahim Alabdulmohsin, et~al.
\newblock Scaling vision transformers to 22 billion parameters.
\newblock In \emph{International Conference on Machine Learning}, pages 7480--7512. PMLR, 2023.

\bibitem[Devlin et~al.(2018)Devlin, Chang, Lee, and Toutanova]{devlin2018bert}
Jacob Devlin, Ming-Wei Chang, Kenton Lee, and Kristina Toutanova.
\newblock Bert: Pre-training of deep bidirectional transformers for language understanding.
\newblock \emph{arXiv preprint arXiv:1810.04805}, 2018.

\bibitem[Dosovitskiy et~al.(2020)Dosovitskiy, Beyer, Kolesnikov, Weissenborn, Zhai, Unterthiner, Dehghani, Minderer, Heigold, Gelly, et~al.]{dosovitskiy2020image}
Alexey Dosovitskiy, Lucas Beyer, Alexander Kolesnikov, Dirk Weissenborn, Xiaohua Zhai, Thomas Unterthiner, Mostafa Dehghani, Matthias Minderer, Georg Heigold, Sylvain Gelly, et~al.
\newblock An image is worth 16x16 words: Transformers for image recognition at scale.
\newblock \emph{arXiv preprint arXiv:2010.11929}, 2020.

\bibitem[Gou et~al.(2021)Gou, Yu, Maybank, and Tao]{gou2021knowledge}
Jianping Gou, Baosheng Yu, Stephen~J Maybank, and Dacheng Tao.
\newblock Knowledge distillation: A survey.
\newblock \emph{International Journal of Computer Vision}, 129\penalty0 (6):\penalty0 1789--1819, 2021.

\bibitem[Hinton et~al.(2015)Hinton, Vinyals, and Dean]{hinton2015distilling}
Geoffrey Hinton, Oriol Vinyals, and Jeff Dean.
\newblock Distilling the knowledge in a neural network.
\newblock \emph{arXiv preprint arXiv:1503.02531}, 2015.

\bibitem[Pei et~al.(2023)Pei, Liu, Li, Shao, Xu, Dai, Lu, and Yan]{pei2023clipping}
Renjing Pei, Jianzhuang Liu, Weimian Li, Bin Shao, Songcen Xu, Peng Dai, Juwei Lu, and Youliang Yan.
\newblock Clipping: Distilling clip-based models with a student base for video-language retrieval.
\newblock In \emph{Proceedings of the IEEE/CVF Conference on Computer Vision and Pattern Recognition}, pages 18983--18992, 2023.

\bibitem[Radford et~al.(2021)Radford, Kim, Hallacy, Ramesh, Goh, Agarwal, Sastry, Askell, Mishkin, Clark, et~al.]{radford2021learning}
Alec Radford, Jong~Wook Kim, Chris Hallacy, Aditya Ramesh, Gabriel Goh, Sandhini Agarwal, Girish Sastry, Amanda Askell, Pamela Mishkin, Jack Clark, et~al.
\newblock Learning transferable visual models from natural language supervision.
\newblock In \emph{International conference on machine learning}, pages 8748--8763. PMLR, 2021.

\bibitem[Wang et~al.(2022)Wang, Codella, Chen, Zhou, Yang, Dai, Xiao, You, Chang, and Yuan]{wang2022clip}
Zhecan Wang, Noel Codella, Yen-Chun Chen, Luowei Zhou, Jianwei Yang, Xiyang Dai, Bin Xiao, Haoxuan You, Shih-Fu Chang, and Lu Yuan.
\newblock Clip-td: Clip targeted distillation for vision-language tasks.
\newblock \emph{arXiv preprint arXiv:2201.05729}, 2022.

\bibitem[Wasim et~al.(2023)Wasim, Naseer, Khan, Khan, and Shah]{wasim2023vita}
Syed~Talal Wasim, Muzammal Naseer, Salman Khan, Fahad~Shahbaz Khan, and Mubarak Shah.
\newblock Vita-clip: Video and text adaptive clip via multimodal prompting.
\newblock In \emph{Proceedings of the IEEE/CVF Conference on Computer Vision and Pattern Recognition}, pages 23034--23044, 2023.

\bibitem[Yang et~al.(2023)Yang, An, Huang, Bi, Yu, Yang, and Xu]{yang2023clip}
Chuanguang Yang, Zhulin An, Libo Huang, Junyu Bi, Xinqiang Yu, Han Yang, and Yongjun Xu.
\newblock Clip-kd: An empirical study of distilling clip models.
\newblock \emph{arXiv preprint arXiv:2307.12732}, 2023.

\bibitem[Zhang et~al.(2022)Zhang, Guo, Zhang, Li, Miao, Cui, Qiao, Gao, and Li]{zhang2022pointclip}
Renrui Zhang, Ziyu Guo, Wei Zhang, Kunchang Li, Xupeng Miao, Bin Cui, Yu Qiao, Peng Gao, and Hongsheng Li.
\newblock Pointclip: Point cloud understanding by clip.
\newblock In \emph{Proceedings of the IEEE/CVF conference on computer vision and pattern recognition}, pages 8552--8562, 2022.

\end{thebibliography}
}


\end{document}